\documentclass[letterpaper, 10 pt, conference]{ieeeconf}      
\usepackage[T1]{fontenc}
\usepackage[letterpaper,top=57pt,bottom=43pt,left=48pt,right=48pt]{geometry}
\usepackage{aecompl}
\usepackage{graphicx}

\usepackage{amsthm,amsmath,amssymb,lipsum}
\usepackage{color}
\usepackage{array}
\usepackage{dsfont}
\usepackage{cite}
\usepackage{multirow}
\usepackage{tabularx}
\usepackage{booktabs}
\usepackage{stfloats}
\usepackage{subfigure}
\usepackage{bm}
\usepackage{float}
\usepackage{caption}
\usepackage{algorithmic}
\usepackage{algorithm}
\usepackage{mathrsfs}

\usepackage{subcaption}
\makeatletter
\let\NAT@parse\undefined
\makeatother
\usepackage{hyperref}
\hypersetup{
colorlinks=true,
linkcolor=blue,
citecolor=cyan,
}
\usepackage{url}
\IEEEoverridecommandlockouts                              

\title{\LARGE \bf
EMP: Executable Motion Prior for Humanoid Robot Standing Upper-body Motion Imitation
}

\author{ Haocheng Xu$^*$, Haodong Zhang$^*$, Zhenghan Chen, Rong Xiong
\thanks{*These authors contributed equally.}
\thanks{Haocheng Xu, Haodong Zhang, Zhenghan Chen, and Rong Xiong are with the State Key Laboratory of Industrial Control and Technology, Zhejiang University, Hangzhou 310027, China. Rong Xiong is the corresponding author
        {\tt\small rxiong@zju.edu.cn}}
}

\begin{document}
\newgeometry{top=60pt,bottom=43pt,left=48pt,right=48pt}

\maketitle
\thispagestyle{empty}
\pagestyle{empty}

\begin{abstract}
To support humanoid robots in performing manipulation tasks, it is essential 
to study stable standing while accommodating upper-body motions. However, the 
limited controllable range of humanoid robots in a standing position 
affects the stability of the entire body. Thus we introduce a reinforcement learning based framework for humanoid robots to 
imitate human upper-body motions while maintaining overall stability. 
Our approach begins with 
designing a retargeting network that generates a large-scale upper-body motion dataset for training the reinforcement learning (RL) policy, which enables the humanoid robot to track upper-body motion targets, employing domain randomization for enhanced robustness. To avoid exceeding the robot's execution capability and ensure safety and stability, we propose an Executable Motion Prior (EMP) module, which adjusts the input target movements based on the robot’s current state. This adjustment improves 
standing stability while minimizing changes to motion amplitude. We 
evaluate our framework through simulation and real-world tests, demonstrating its practical applicability. \href{https://anonymous.4open.science/w/EMP-project-page-4D58/}{Project page}.

\end{abstract}

\section{Introduction}

The humanoid form allows humanoid robots to better adapt to human environments, tools, 
and human-machine interactions.
We aim to enable humanoid robots to perform human-like movements, allowing 
for better mapping of human motions onto the robots. This enables them to quickly 
learn human motion skills, which lays the foundation for executing subsequent task operations.

However, many challenges remain in the practical implementation of humanoid robots 
mimicking human motions. The complex dynamic characteristics of humanoid robots, 
along with their high-dimensional state and action spaces, complicate motion control. 
While model-based controllers have shown remarkable results in whole-body motion 
imitation \cite{9158331,8461207,8375643}, the computational burden of complex 
dynamics models restricts these methods to simplified models, limiting their 
scalability for dynamic motions.

Recently, reinforcement learning methods have gained popularity in the field 
of humanoid robotics. Initially, RL was employed in the graphics community to 
generate humanoid motions from human motion data for animated characters \cite{ASE,AMP}.
Additionally, RL controllers have been developed for bipedal robot walking 
\cite{siekmann2021blindbipedalstairtraversal, li2024reinforcementlearningversatiledynamic}, 
whole-body control \cite{cheng2024expressivewholebodycontrolhumanoid}, and 
humanoid teleoperation \cite{he2024learninghumantohumanoidrealtimewholebody}.

\begin{figure}[ht]
    \vspace{0.5em}
    \centering
    \includegraphics[scale=0.3]{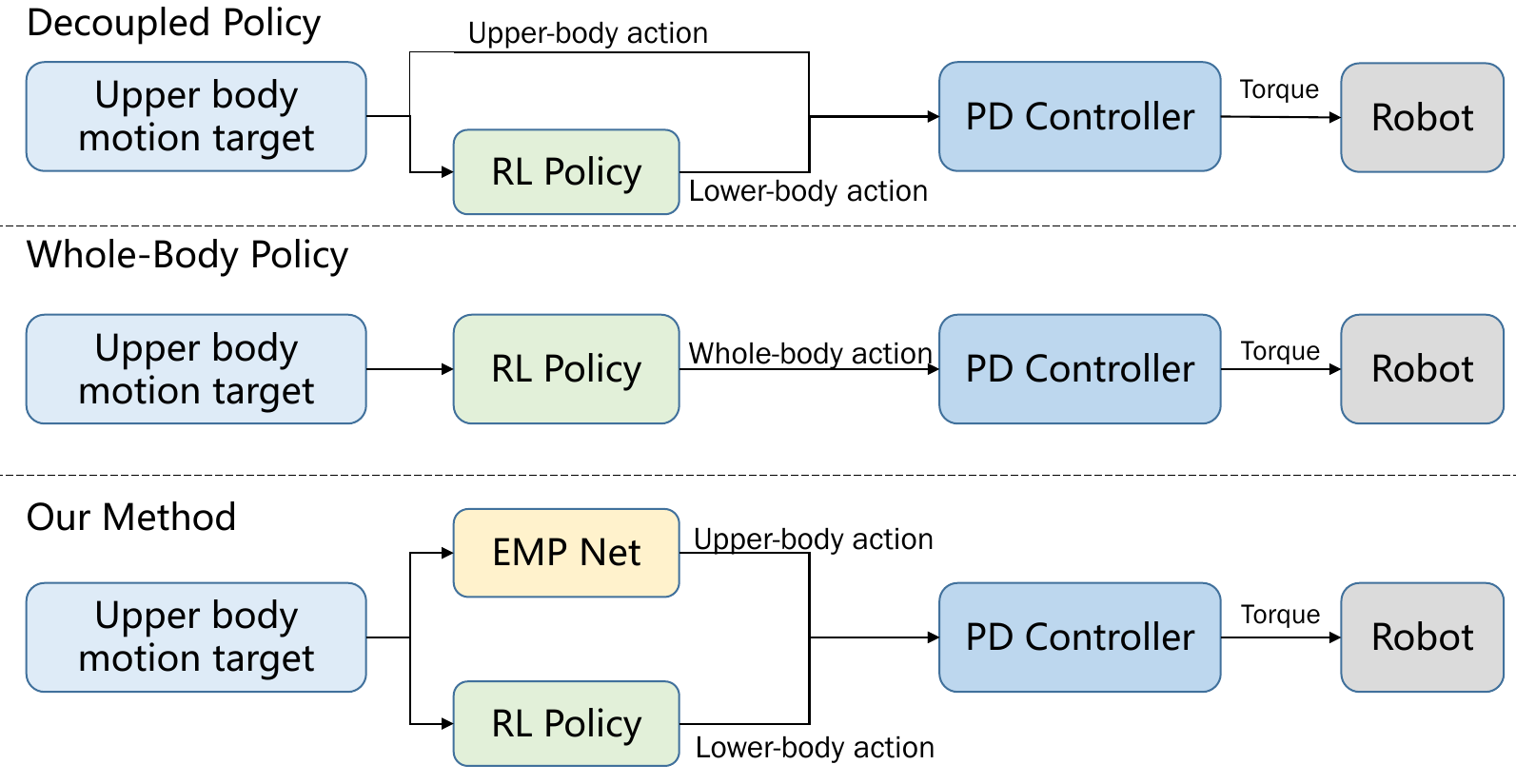}
    \caption{Different Motion Imitation Framework. (a) Decoupled Policy, such as PMP \cite{lu2025mobiletelevisionpredictivemotionpriors}, only generates lower-body actions and execute upper-body target straightly. (b) Whole-Body Policy, like HumanPlus\cite{fu2024humanplus} and Exbody\cite{cheng2024expressivewholebodycontrolhumanoid}, controls whole body joint to imitate whole body motion target. (c) Our method introduce a executable motion prior to optimize upper-body motion target while RL policy provides lower-body actions.}
    \label{intro}
    \vspace{-2.0em}
\end{figure}

Our work focuses on humanoid robots imitating human upper-body movements 
because humanoid robots perform tasks with their upper bodies while standing most of time. The mainstream frameworks are shown in Figure \ref{intro}.
In humanoid reinforcement learning for motion imitation, we observe a conflict between 
stability and similarity rewards. When joints are entirely controlled by the RL 
policy for whole-body control like \cite{he2024learninghumantohumanoidrealtimewholebody,fu2024humanplus}, vibration and deviation on base and upper-body actions can 
occur. Conversely, directly executing upper-body actions may 
lead to the robot's limited control capacity exceeding the RL policy's capabilities, 
resulting in a loss of balance.

\begin{figure*}[ht]
    \vspace{1.0em}
    \begin{center}
        \centering
        \includegraphics[scale=0.53]{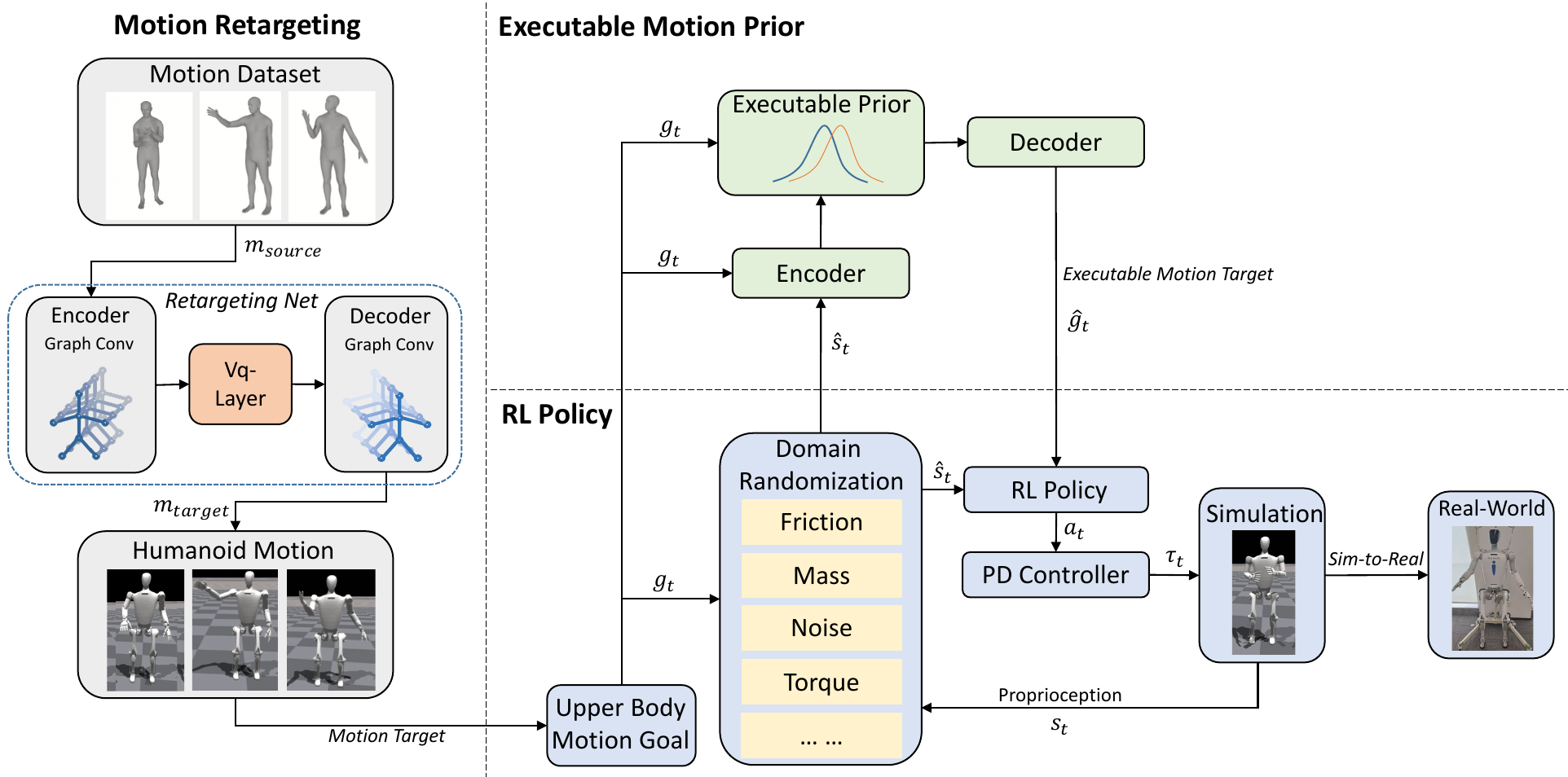}
    \end{center}
    \caption{Overview of our framework. 
    \textbf{Motion Retargeting} (section \ref{Retargeting Human Motion to Humanoids}): We train a graph convolution retargeting network to convert human motions $\boldsymbol{m}_{source}$ to humanoid joint actions $\boldsymbol{m}_{target}$ as motion goal for imitation. 
    \textbf{RL Policy} (section \ref{RL}): We train an upper-body imitation policy for the humanoid to track the upper-body motion goal $\boldsymbol{g}_t$ while keeping balance.
    \textbf{Executable Motion Prior} (section \ref{Executable Motion Prior}): We use a VAE-based network to adjust the goal motion based on the current state $\hat{\boldsymbol{s}}_t$, improving stability.
    }
    \label{overview}
    \vspace{-1.0em}
\end{figure*}

In this paper, we present a system for humanoid robots to imitate human upper-body motions 
while maintaining whole-body stability. Combined with imitation learning and reinforcement learning,
Figure \ref{overview} shows our framework.
First, we design a graph convolutional 
network to retarget human motions to humanoid movements, creating a motion dataset for 
training a robust RL imitation policy. Next, we train a RL policy for upper-body 
motion imitation using retargeted motions. This policy manages the lower-body 
joints to maintain balance, while upper-body targets are directly sent to robot to ensure alignment with targets.

When humans are performing upper-body actions, they can recognize potential dangers and make motion adjustments in a timely manner. Inspired by this, we propose an \textbf{E}xecutable \textbf{M}otion \textbf{P}rior 
(\textbf{EMP}) that modifies the input target upper-body motions based on the 
robot's current state. This approach enhances standing stability while minimizing 
alterations to the motion amplitude. Utilizing the dataset obtained from motion 
retargeting and the trained RL controller, we train an EMP network. 
This network transforms unstable actions into stable ones by simultaneously encoding 
the robot's current state and action objectives into a latent space and decoding them 
into new, more reasonable action objectives, functioning as an action optimization 
module before RL controller. Finally we deploy this framework in real-world humanoid robots.

Our contributions are as follows:
\begin{itemize}
    \item[1)] A RL based framework for humanoid imitating upper-body motion, which includes a motion retargeting network to transfer human motions to humanoid motions and a RL policy to control the robot while tracking upper-body motions;
    \item[2)] An executable motion prior for the RL imitation policy system that adjusts target motions based on the humanoid's current state, enhancing stability while minimizing changes in motion amplitude;
    \item[3)] A world model to simulate the state transition process of the environment for gradient backpropagation.
    \item[4)] Sim-to-real transfer of our system that demonstrates its effectiveness in two humanoid robots.
\end{itemize}

\section{Related Works}

\subsection{Motion Retargeting}
Motion retargeting facilitates the transfer of motion data from a source character to 
a target character. In the context of animation, both optimization-based methods 
\cite{10.1145/311535.311536,rekik2024correspondencefreeonlinehumanmotion} 
and learning-based methods \cite{10.1145/2897824.2925975,10.1145/3283289.3283316} 
are employed for motion transfer between animated characters. 
Learning-based methods tend to yield more efficient results and can facilitate 
motion transfer across different skeletal structures \cite{Aberman_2020,10496240}.

In humanoid motion retargeting, Delhaisse et al. \cite{7966379} use shared latent
variable models to retarget motions between different humanoids. 
Ayusawa et al. \cite{8063558} reconstruct human motion within the physical 
constraints imposed by humanoid dynamics and offer precise morphing function for 
different human body dimensions. In our work we design a network to 
generate humanoid motions from human motions.

\subsection{Reinforcement Learning for Humanoid }
Before RL-based controllers being used into real-world humanoids, they are often used
in physics-based animation control \cite{luo2023perpetualhumanoidcontrolrealtime,ASE,AMP}.
Nevertheless, the humanoid avatars usually have a high degree of freedom, 
with minimal restrictions on joint positions, torques, and sometimes with additional auxiliary force \cite{10.5555/3495724.3497550}.

On the other hand, realistic humanoids have complex dynamic models, and it is difficult to 
obtain privileged states, such as base velocity and height, from build-in sensors \cite{cheng2024expressivewholebodycontrolhumanoid}.
This makes it impossible to directly transfer RL models 
used for animated characters to physical humanoids.  Li et al. \cite{li2024reinforcementlearningversatiledynamic}
proposed an end-to-end RL approach and use task randomization to build a robust
dynamic locomotion controller for bipedal robots. Radosavovic et al. \cite{HumanoidTransformer2023} designed
causal Transformer trained by autoregressive prediction of future actions from the history of observations and actions
for real-world full-sized humanoid locomotion. Siekmann et al. \cite{siekmann2021blindbipedalstairtraversal}
use stair-like terrain randomization to build a RL controller for humanoid traversing stair-like terrain.
In this work we build a sim-to-real training process with domain randomization and 
deploy our policy to realistic humanoid.

\subsection{Humanoids Imitation from Human Motion}
Traditional methods, such as model predictive control (MPC),
use model-based optimization methods to minimize tracking errors under stability and contact constraints \cite{10035484}. 
However, due to the high computational burden, the humanoid model is usually simplified \cite{zhang2023slomogeneralleggedrobot,doi:10.1126/scirobotics.aav4282},
which limits the accuracy of imitation. 

RL controllers provides an alternative solution. Cheng et al. \cite{cheng2024expressivewholebodycontrolhumanoid}
train a whole-body humanoid controller with a large-scale motion dataset.
He et al. \cite{he2024learninghumantohumanoidrealtimewholebody}
use a privileged policy to select a executable motion dataset, which helps training
a robust RL policy for sim-to-real deployment. Fu et al. \cite{fu2024humanplus}
train a task-agnostic low-level policy to track retargeted humanoid poses. Lu et al. \cite{lu2025mobiletelevisionpredictivemotionpriors} proposed a CVAE-based motion prior to enhance the robustness of controller.
In this work we focus on tracking the upper-body motion targets and build a executable
motion prior network to filter motions and a upper-body motion imitation RL policy to
keep whole-body balance while tracking upper-body targets.

\section{Retargeting Human Motion to Humanoids}
\label{Retargeting Human Motion to Humanoids}

\subsection{Retargeting Network Architecture}
Prior work \cite{10496240} has achieved cross-skeleton motion retargeting between animated characters with graph network.
We develop a network for motion retargeting from human to humanoid. Using networks for motion retargeting offers good real-time performance and generalization capabilities.

Figure \ref{overview} illustrates the structure of our retargeting net. 
We regard the upper-body skeleton of the humanoid and the human as a graph.
Referring to the framework of VQ-VAE \cite{oord2018neuraldiscreterepresentationlearning}, 
our network consists of a motion encoder, a vq-codebook layer and a motion decoder. 

The motion encoder $f_e$ embeds the source motion from human. The source motion is represented 
as the positions of key nodes $\bm{Q}_A \in \mathbb{R}^{N_A \times 3} $ 
and the features of edges $\bm{E}_A $. After passing through the graph convolutional layers, 
the source motion features are encoded into the latent space features :
    $\bm{z}_A = f_e(\bm{Q}_A,\bm{E}_A)$.
A transformation net $f_{tf} $ converts the latent features of input skeleton A 
into the latent features of output skeleton B:
    $\bm{z}_B = f_{tf}(\bm{z}_A)$.
Then the codebook layer choose the nearest element of latent embedding vectors:
\begin{equation}
    \bm{z}_e = \bm{e}_k   \mbox{  where  } k=\arg\min_j \Vert \bm{z}_B - \bm{e}_j \Vert_2
\end{equation}

The motion decoder $f_d$ generates the target motion $\bm{Q}_B\in \mathbb{R}^{N_B}$ (represented by joint angles) 
with latent embedding vector $\bm{z}_e$ and edge features $\bm{E}_B $:
    $\bm{Q}_B = f_d(\bm{z}_e,\bm{E}_B)$.

The key nodes are waist, torso, shoulder, elbow and wrist.

\begin{table}[h]

    \renewcommand{\arraystretch}{1.25}
    \begin{center}
        \caption{Training Loss for Retargeting Network}
\label{Training Loss}
\begin{tabular}{lcr}
\toprule[0.7pt]
\multicolumn{1}{l}{Term} & \multicolumn{1}{c}{Expression} & \multicolumn{1}{r}{Weight}\\
\midrule[0.5pt]
$L_{ee}$        &    $\displaystyle \Vert \frac{\bm{p}^{ee} - \bm{p}^{elb}}{\Vert\bm{p}^{ee} - \bm{p}^{elb}\Vert_2} - \frac{\hat{\bm{p}}^{ee} - \hat{\bm{p}}^{elb}}{\Vert\hat{\bm{p}}^{ee} - \hat{\bm{p}}^{elb}\Vert_2} \Vert_2^2$       &  100   \\[2ex]
\midrule[0pt]
$L_{ori}$        &    $\displaystyle \Vert \bm{R} - \hat{\bm{R}} \Vert_2^2$       &  100   \\[2ex]
\midrule[0pt]
$L_{elb}$        &    $\displaystyle \Vert \frac{\bm{p}^{elb} - \bm{p}^{sho}}{\Vert\bm{p}^{elb} - \bm{p}^{sho}\Vert_2} - \frac{\hat{\bm{p}}^{elb} - \hat{\bm{p}}^{sho}}{\Vert\hat{\bm{p}}^{elb} - \hat{\bm{p}}^{sho}\Vert_2} \Vert_2^2$        &  100   \\[2ex]
\midrule[0pt]
$L_{emb}$        &    $\displaystyle \Vert \mbox{sg}(\bm{z}_e)-\bm{e}\Vert_2^2$       &  10000   \\[2ex]
\midrule[0pt]
$L_{com}$        &    $\displaystyle 0.25\Vert \bm{z}_e-\mbox{sg}(\bm{e})\Vert_2^2$      &  10000   \\[2ex]

\toprule[0.7pt]

\end{tabular}
\end{center}
\vspace{-1.0em}
\end{table}

\subsection{Training Loss}
Combined with the method in \cite{zhang2022kinematicmotionretargetingneural}, 
the training loss of our retargeting network is composed of
five terms: end effector loss $L_{ee}$, orientation loss $L_{ori}$, elbow loss $L_{elb}$, 
embedding loss $L_{emb}$ and commitment loss $L_{com}$. 
We list the losses in Tab \ref{Training Loss}, where $\bm{p}$ and $\hat{\bm{p}}$ mean the 
node position of human and humanoid respectively, $\bm{R}$ and $\hat{\bm{R}}$ mean the end
effector (namely wrist) rotation matrix, sg() means stop gradient.

\begin{table}[h]
    \vspace{0.5em}
    \renewcommand{\arraystretch}{1.25}
    \begin{center}
        \caption{Rewards Expressions and Weights}
\label{reward}
\begin{tabular}{ccc}
\toprule[0.7pt]
\multicolumn{1}{l}{Term} & \multicolumn{1}{l}{Expression} & \multicolumn{1}{r}{Weight}\\
\hline
& \multicolumn{1}{l}{Regularization}            &\\
\hline
\multicolumn{1}{l}{Base orientation}        & \multicolumn{1}{l}{$ \exp{(-10\Vert \bm{rpy}_t^{xy} \Vert_1)} $}           &  \multicolumn{1}{r}{3.0}   \\
\multicolumn{1}{l}{Projected gravity}        & \multicolumn{1}{l}{$\exp{(-20\Vert \bm{pg}_t^{xy} \Vert_2)} $}           &  \multicolumn{1}{r}{3.0}   \\

\multicolumn{1}{l}{Base height}        & \multicolumn{1}{l}{$\exp{(-100|h_t - h^{\mbox{ref}}|)}$}           &  \multicolumn{1}{r}{0.2}   \\
\multicolumn{1}{l}{Base linear velocity}        & \multicolumn{1}{l}{$\exp{(-10\Vert \bm{v}_t \Vert_2^2)} $}           &  \multicolumn{1}{r}{0.75}   \\
\multicolumn{1}{l}{Base angular velocity}        & \multicolumn{1}{l}{$ \exp{(-20\Vert \bm{\omega_t}\Vert_2)} $}           &  \multicolumn{1}{r}{0.75}   \\
\multicolumn{1}{l}{Base acceleration}        & \multicolumn{1}{l}{$\exp{(-3\Vert \bm{v}_t - \bm{v}_{t-1} \Vert_2)}$}           &  \multicolumn{1}{r}{0.2}   \\

\multicolumn{1}{l}{Leg DoF position}        & \multicolumn{1}{l}{$\exp{(-100\Vert \bm{q}_t^{\mbox{leg}} - \bm{q}^{\mbox{leg,ref}} \Vert_2)}$}           &  \multicolumn{1}{r}{1.0}   \\
\multicolumn{1}{l}{Feet contact}        & \multicolumn{1}{l}{$ \mathds{1}(F_{\mbox{feet}}^z \geqslant 5) $}                             &  \multicolumn{1}{r}{0.5}   \\
\multicolumn{1}{l}{Feet slip}           & \multicolumn{1}{l}{$ \mathds{1}(F_{\mbox{feet}}^z \geqslant 5) \times \sqrt{\Vert \bm{v}_t^{\mbox{feet}} \Vert_2} $}           &  \multicolumn{1}{r}{0.2}   \\

\hline
                                        & \multicolumn{1}{l}{Energy}                    &\\
\hline
\multicolumn{1}{l}{Action range}   & \multicolumn{1}{l}{$ \Vert \bm{a}_t \Vert_1   $}                   & \multicolumn{1}{r}{-0.075} \\
\multicolumn{1}{l}{Action rate}         & \multicolumn{1}{l}{$\Vert \bm{a}_t - \bm{a}_{t-1} \Vert_2^2$}               &  \multicolumn{1}{r}{-1.5} \\
\multicolumn{1}{l}{Action acceleration} & \multicolumn{1}{l}{$\Vert \bm{a}_t + \bm{a}_{t-2} - 2\bm{a}_{t-1} \Vert_2^2$}  &  \multicolumn{1}{r}{-1.5} \\
\multicolumn{1}{l}{Torques}             & \multicolumn{1}{l}{$\Vert \bm{\tau}_t \Vert_2^2$ }                     & \multicolumn{1}{r}{-1e-5} \\
\multicolumn{1}{l}{Dof velocity}        & \multicolumn{1}{l}{$\Vert \dot{\bm{q}_t} \Vert_2^2$  }                &  \multicolumn{1}{r}{-1e-4} \\
\multicolumn{1}{l}{Dof acceleration}    & \multicolumn{1}{l}{$\Vert \ddot{\bm{q}_t} \Vert_2^2$  }                &  \multicolumn{1}{r}{-1e-7} \\
\toprule[0.7pt]

\end{tabular}
\end{center}
\vspace{-1.0em}
\end{table}

\section{RL Control Policy Training for Humanoid Upper-Body Imitation}
\label{RL}

\subsection{Overview}
We decouple the whole-body control policy into $\pi_{lower}$ and $\pi_{upper}$. $\pi_{lower}$ is an RL-based policy which generates lower-body actions from proprioception state to keep the humanoid robot standing in balance while tracking upper-body motions. The upper-body policy $\pi_{upper}$ is a open loop controller, namely our executable motion prior (EMP) network, which is detailed in Section \ref{Executable Motion Prior}.

\subsection{State Space}
\label{state space}
We consider our RL control policy as a goal-conditioned policy 
$\pi: \textbf{G} \times \textbf{S} \longrightarrow \textbf{A}$, 
where $\textbf{G}$ is goal space that indicates the upper-body motion target , 
$\textbf{S}$ is the observation space and $\textbf{A}$ is the action space for lower-body joints.

We define goal state as  $ \bm{g}_t \triangleq \bm{q}_{\mbox{target}} \in \mathbb{R}^{15}$, 
where $\bm{q}_{\mbox{target}}$ represents the target joint position of upper-body joints, 
including two 7-dof arms and one 1-dof waist. The action is denoted as $\bm{a}_t \in \mathbb{R}^{12}$
. We define our observation state as $\bm{s}_t \triangleq [\bm{q}_t,\bm{a}_{t-1}, \bm{rpy}_t, \bm{g}_t]$,
where  $\bm{q}_t \in \mathbb{R}^{27} $ indicates the joint position 
and $\bm{rpy}_t \in \mathbb{R}^{3}$ is the euler angle of robot base. We combine states of last $T$ frames together
as $\bm{S}_t = \{\bm{s}_{t-T:t}\} \in \mathbb{R}^{T \times 65}$ to utilize history message. We set 
$T = 15$ in experiments. 
The action space consists of 12-dim joint position targets (two 6-dof legs).
The joint actions will be converted to joint torque by a PD controller.

\subsection{Reward Design}
The rewards are detailed
in Tab \textcolor{blue}{\ref{reward}},
where $h^{\mbox{ref}}$ is reference height of base, 
$\bm{q}^{\mbox{leg,ref}} $ is reference joint positions of legs. 

Our policy focuses on upper-body motion imitation while standing, 
so we just set $\bm{v}^{\mbox{ref}}_t=0$ and $\bm{\omega}^{\mbox{ref}}_t=0$.

\begin{table}[h]
    \renewcommand{\arraystretch}{1.25}
    \begin{center}
        \caption{Domain Randomization}
\label{Domain}
\begin{tabular}{cc}
\toprule[0.7pt]
Term  &   Value \\
\hline
Friction   &   $\mathcal{U}(0.1,2.0)$ \\
Base Mass   &  $\mathcal{U}(-5.0,5.0) + \mbox{default kg}$ \\
Hand Mass   &  $\mathcal{U}(0,2.5) + \mbox{default kg}$ \\
Base Com   &   $\mathcal{U}(-0.05,0.05)$ m \\
Link Inertia &  $\mathcal{U}(0.8,1.2) \times \mbox{default kg}\cdot \mbox{m}^2$  \\
Link Mass &  $\mathcal{U}(0.8,1.2) \times \mbox{default kg}$ \\
P Gain  &    $\mathcal{U}(0.8,1.2) \times \mbox{default}$ \\
D Gain  &    $\mathcal{U}(0.8,1.2) \times \mbox{default}$ \\
Motor Torque   &  $\mathcal{U}(0.8,1.2) \times \mbox{default } \mbox{N} \cdot \mbox{m}$ \\
Motor Damping   &  $\mathcal{U}(0.3,4.0)  \mbox{ N} \cdot \mbox{s}$ \\
Motor Delay   &  $\mathcal{U}(0,10) \mbox{ ms}$ \\
Push Robots   &  interval = 5s, $v_{xy}=0.5 \mbox{m/s}, \omega=0.4 \mbox{rad/s}$  \\
Hang Robots    &  height = 0.1m, ratio = 20\%   \\
Init Joint Position    &  $\mathcal{U}(-0.1,0.1)+ \mbox{default}$ rad  \\
Action      &  $\mathcal{U}(0.98,1.02) \times \mbox{default}$ \\

\toprule[0.7pt]
\end{tabular}
\end{center}
\vspace{-1.0em}
\end{table}

\subsection{Domain Randomization}

The domain randomization we use in our policy are listed in
Tab \ref{Domain}. 
we add random mass to the hands separately to enhance the terminal load capacity 
and We raise the robot by 0.1m with a probability of 20\% during initialization.

\subsection{Termination Conditions}
\label{Termination}
To improve training efficiency, we reset training process when the
projected gravity on x or y axis exceeds 0.7.

\begin{figure}[ht]
    \vspace{0.5em}
    \centering
    \includegraphics[scale=0.33]{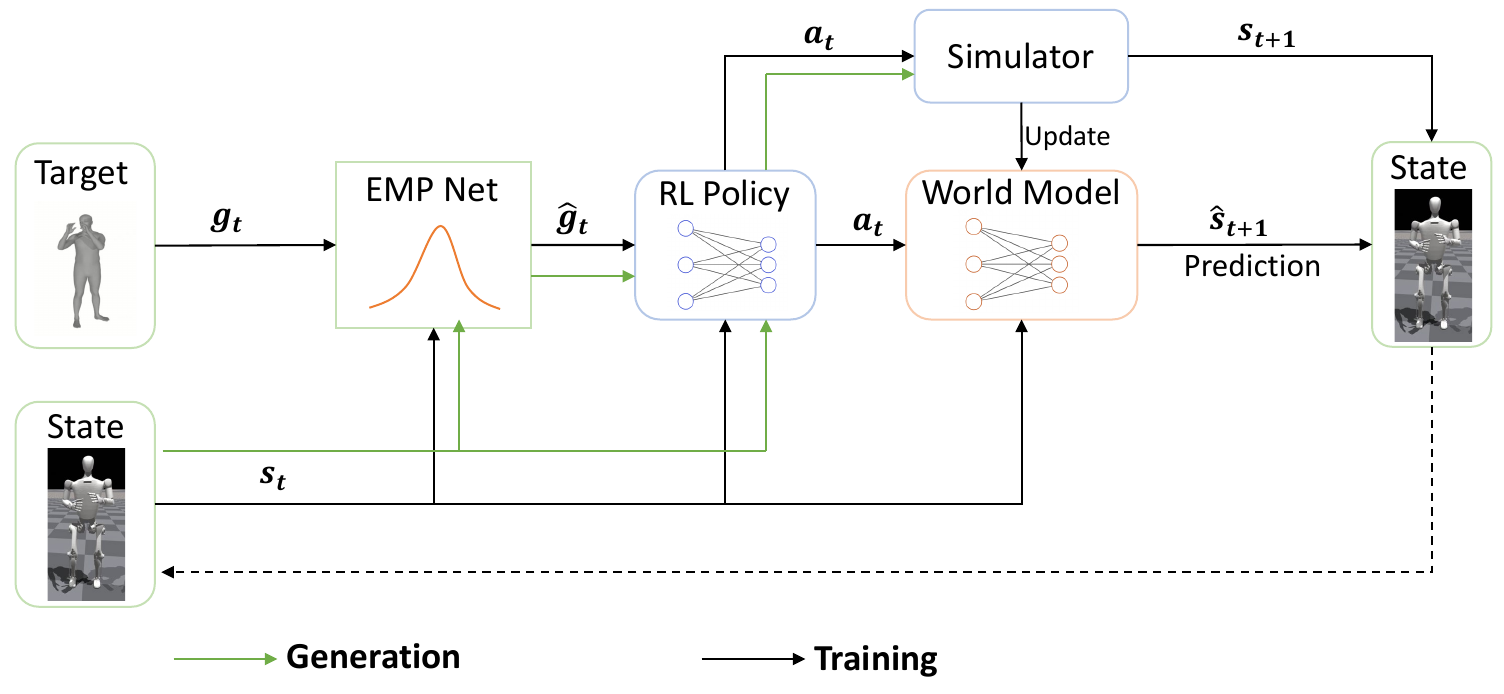}
    \caption{Framework of our EMP System. EMP network generates optimized upper-body motion targets conditioned on the state of the robot. The world model learns the state transition model from the simulator for gradient backpropagation. }
    \label{EMP}
    \vspace{-1.5em}
\end{figure}

\section{Executable Motion Prior }
\label{Executable Motion Prior}

\subsection{Overview}

Our executable motion prior (EMP) network is designed based on the structure of 
Variational Autoencoder (VAE) \cite{kingma2022autoencodingvariationalbayes}. 
Inspired by the framework of ControlVAE\cite{shao2020controlvaecontrollablevariationalautoencoder}, 
we build our overall framework, shown in Figure \ref{EMP}.
EMP adjusts the target upper-body motion based on the current state of the humanoid, improving the humanoid's 
standing stability while minimizing changes in the motion amplitude.

The EMP network consists of an encoder and a decoder, showed in Figure \ref{overview}. 
The encoder is composed of 3 sub-networks: a state encoder $f_s$, a target encoder $f_t$ and a fusion network $f_\psi$. The state encoder and target encoder encode the state and the target into the latent space variable $\bm{z}_1,\bm{z}_2$, respectively. 
\begin{equation}
    \bm{z}_1,\bm{z}_2 = f_s(\bm{s}_t),f_t(\bm{g}_t)
\end{equation}

Then the fusion network encodes two variables into a single latent space vector $\bm{z} = f_{\psi}(\bm{z}_1,\bm{z}_2)$, 
which follows a standard normal distribution $\bm{z}\thicksim \mathcal{N}(0,1) $
and then the decoder generates new target for the humanoid. Then the EMP 
can be described as:
\begin{equation}
    \hat{\bm{g}}_t = f_{\theta}(\bm{s}_t,\bm{g}_t)
\end{equation}
where $\theta$ is the learnable variable of EMP net. The encoder and decoder net are both MLP networks.

\subsection{Training}
The training process consists of two parts: world model $f_w$ training and EMP training.

\noindent \textbf{World Model Training}. Due to the inability to obtain gradients 
from the robot state information in the simulation,
we use a world model to simulate the state transition process of the humanoid robot environment.
The world model predicts the next state of the humanoid robot depending on the current state and action:
\begin{equation}
    \hat{\bm{s}}_{t+1} = f_w(\bm{s}_t,\bm{a}_t)
\end{equation} 
where $w$ is the learnable variable of world model. Then we have world model prediction loss:
\begin{equation}
    L_{pre} = \Vert\bm{s}_{t+1} - \hat{\bm{s}}_{t+1}\Vert_2^2
\end{equation}
where $\bm{s}_{t+1}$ is the state given by the simulator, namely isaacgym here 
and $\hat{\bm{s}}_{t+1}$ is the prediction of the world model. 
The state here is defined the same as section \ref{state space}. 

\noindent \textbf{EMP Training}.
While the robot is losing its balance, the following situations usually occur: (1) The center of gravity is projected away from the support surface; (2) The robot's torso is no longer oriented vertically upwards. Therefore we train the network to avoid these situations. Meanwhile, the self-collision and smoothness of the motion can also influence the balance.

The training process of EMP is illustrated in Figure \ref{EMP}. We have the following losses:

\noindent i) \textit{Reconstruction Loss}. The reconstruction loss $L_{rec}$ encourages 
the generated motion $\hat{\bm{g}}_t$ to be as identical to the source target $\bm{g}_t$. We define
\begin{equation}
    L_{rec} = \Vert \bm{g}_t -  \hat{\bm{g}}_t \Vert_2^2
\end{equation}

\noindent ii) \textit{Orientation Loss}. The orientation loss  $L_{ori}$ promotes the humanoid's base
to stay upright, which can improve the stability of the humanoid. Then $L_{ori}$ is defined as
\begin{equation}
    L_{ori} =\exp(-\Vert \widehat{\bm{pg}}_{t+1}^{xy} \Vert_2^2)-1
\end{equation}
where $\widehat{\bm{pg}}_{t+1}^{xy}$ is the projected gravity vector, which is calculated from
$\widehat{\bm{rpy}}_{t+1}^{xy}$ predicted by the world model. 

\noindent iii) \textit{Collision Loss}. The collision loss $L_{col}$ encourages the motion 
to reduce self-collision of the humanoid. We simplify the links that may collide into a spherical model,
and calculate the distance between the links. We define
\begin{equation}
    L_{col} = \sum_{i,j \in \mathbb{J} } \exp[-2(0.08 - \Vert \bm{p}_i-\bm{p}_j \Vert_2)]
\end{equation}
where $\mathbb{J}$ is the set of the links that may collide each other, we define
$\mathbb{J} = \{\mbox{torso, hand, sacrum, thigh} \}$ here. $\bm{p}_i$ and $\bm{p}_j$
mean the coordinate of the links centers, which can be calculated with forward kinematics (FK). 

\noindent iv) \textit{Centroid Loss}. The centroid loss $L_{cen}$ prompts the centroid of humanoid
to stay in the range of support surface under foot. $L_{cen}$ is defined as
\begin{equation}
    L_{cen} = \min\{\exp(-7(0.03-d)), 10\}-1
\end{equation}
where $d$ is the distance between the center of the foot support surface and the projection 
of the centroid onto the ground.

\noindent v) \textit{Smoothness Loss}.  The smoothness loss $L_{smo}$ promotes the motion to
be smooth and reduce the occurrence of motion mutations. $L_{smo}$ is defined as
\begin{align} 
    L_{smo} = \Vert \hat{\bm{g}}_t -  \hat{\bm{g}}_{t-1}  \Vert_2^2 + 0.2\Vert \hat{\bm{g}}_t + \hat{\bm{g}}_{t-2} -  2\hat{\bm{g}}_{t-1}  \Vert_2^2
\end{align}

\noindent vi) \textit{Regularization Loss}. The regularization loss $L_{reg}$ encourages
the latent variable to conform to standard Gaussian distribution. $L_{reg}$ is defined as
\begin{equation}
    L_{reg} = \Vert \bm{z} \Vert_2^2
\end{equation}
where $\bm{z}$ is the latent variable.

Finally we get overall loss for EMP training:
\begin{align}
    L = &\lambda_{rec}L_{rec} + \lambda_{ori}L_{ori} + \lambda_{col}L_{col} \notag \\ 
        &+ \lambda_{cen}L_{cen} + \lambda_{smo}L_{smo} + \lambda_{reg}L_{reg}
\end{align}
We set $\lambda_{rec}=20, \lambda_{ori}=10, \lambda_{col}=1, \lambda_{cen}=10, \lambda_{smo}=100, \lambda_{reg}=1$ here. 
The overall training process is shown in Algorithm \ref{Training process of EMP}.

\begin{algorithm}[ht]
    \caption{Training process of EMP}
    \label{Training process of EMP}
    \begin{algorithmic}[1]
        \STATE \textbf{for} number of training epochs \textbf{do}:
        \STATE \quad \textbf{for} batch of motions in training set \textbf{do}:
        \STATE \qquad Reset simulation environment;
        \STATE \qquad \textbf{for} $t \leftarrow 0 \mbox{ \textbf{to} } T-1$ \textbf{do}:
        \STATE \quad \qquad Sample $\bm{a}_t = \pi(\bm{s}_t,\bm{g}_t)$;
        \STATE \quad \qquad Sample $\bm{s}_{t+1}$ and $\hat{\bm{s}}_{t+1} = f_w(\bm{s}_t,\bm{a}_t)$;
        \STATE \quad \qquad Update world model $f_w$ with $\nabla_w L_{pre}$;
        \STATE \quad \quad \textbf{end for}
        \STATE \qquad Reset simulation environment;
        \STATE \quad \quad \textbf{for} $t \leftarrow 0 \mbox{ \textbf{to} } T-1$ \textbf{do}:
        \STATE \quad \qquad Sample $\hat{\bm{g}}_t = f_{\theta}(\bm{s}_t,\bm{g}_t)$;
        \STATE \quad \qquad Sample $\bm{a}_t = \pi(\bm{s}_t,\hat{\bm{g}}_t )$;
        \STATE \quad \qquad Sample $\hat{\bm{s}}_{t+1} = f_w(\bm{s}_t,\bm{a}_t)$;
        \STATE \quad \qquad Update EMP $f_{\theta}$ with $\nabla_{\theta} L$;
        \STATE \qquad \textbf{end for}
        \STATE \quad \textbf{end for}
        \STATE \textbf{end for}
    \end{algorithmic}
\end{algorithm}

\subsection{Generation}
The Generation process is illustrated in Figure \ref{EMP}. 
With the trained prior distribution, 
the EMP net generates the executable target for humanoid from source target 
and state.

\section{Experiments}

\subsection{Simulation Experiments}
\label{simulation}

\begin{figure*}[ht]
    \centering
    \includegraphics[scale=0.5]{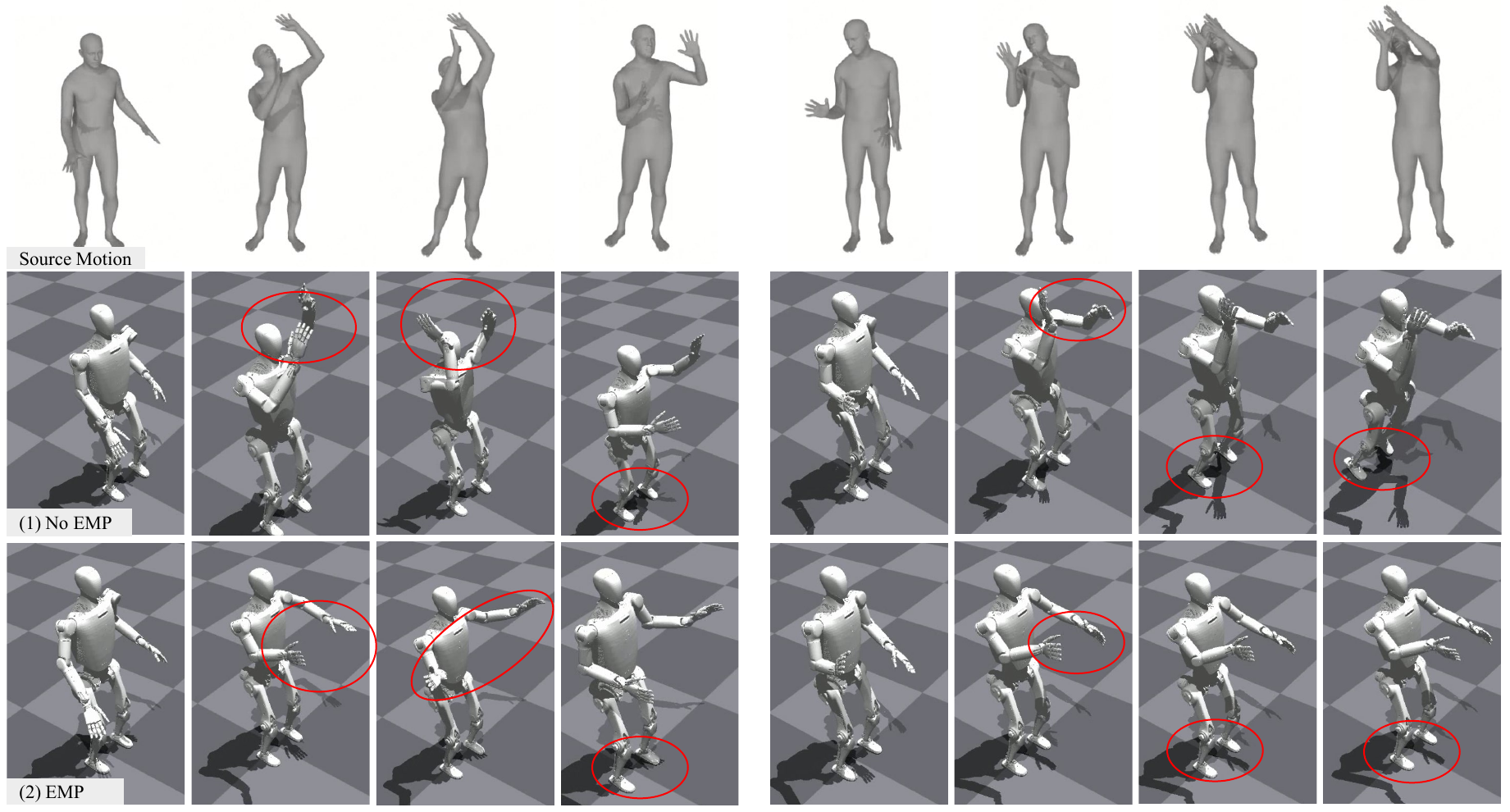}
    \caption{Simulation experiments (left motion: hammer use, right motion: lightbulb screw). The results show that while executing dangerous motions, EMP network will optimize the unexecutable motion and keep the robot stand stably.}
    \label{hi sim}
    \vspace{-1.5em}
\end{figure*}

\noindent \textbf{Hardware Platform}. The main humanoid platform we use is a full-sized robot (1.65m, 60kg) which 
feature 27 degrees of freedom, including two 7-dof arms (about 6kg for one arm, which brings higher load capacity and control difficulty) , two 6-dof legs and one 1-dof in waist.

\noindent \textbf{Implementation Details}. The encoder and decoder of retargeting network are both graph convolutional neural network with three graph convolutional layers, and the hidden sizes are [16,32,64] and [66,32,16], respectively. The codebook of retargeting network has 2048 latent space vectors, each with a dimensionality of 64. The world model is implemented as multi-layer
perceptrons (MLP) with hidden size of [1024,512]. The state encoder and target encoder of EMP network are MLPs with hidden sizes of [1024,1024], and the fusion network and decoder are MLPs with hidden sizes of [2048,2048].
The RL training is conducted on an NVIDIA A800 (80GB) GPU and takes about 6 hours with a learning rate of 1e-3 in Isaac Gym\cite{makoviychuk2021isaacgymhighperformance}. The EMP network is trained on an NVIDIA RTX4060 GPU for 5 hours.

\noindent \textbf{Motion Dataset}.
We use our retargeting network to build our humanoid 
motion dataset. We choose GRAB dataset \cite{Taheri_2020} in AMASS dataset \cite{AMASS:ICCV:2019}
as our source motion dataset. We train our network on the dataset and use the retargeting results for RL policy training.
We divide these motions into smaller motions of the same length, 
with each motion being 60 frames long, and then reconnect them to 
eliminate the inconvenience caused by varying motion lengths. 
For EMP training,
we divide these motions into smaller motions with 200 frames long, facilitating 
our batch collecting process.

\noindent \textbf{EMP Training}. We train our Executable motion prior (EMP) on retargeted GRAB dataset, 
which we randomly divided into a training set (1,070 motions) and a test set (270 motions).
We train the world model and EMP with Adam \cite{kingma2017adammethodstochasticoptimization} optimizer with an initial learning rate of 1e-3.

\noindent \textbf{Baselines}. We consider the following baselines:
\begin{itemize}
    \item[\romannumeral1)] \textbf{Privileged Policy}. Referring to the settings in \cite{he2024learninghumantohumanoidrealtimewholebody}, the observation space for the privileged policy input includes all first-hand robot state, 
    and no noise or domain randomization is added during training. 
    The privileged policy demonstrates the upper limit of the robot's mobility.
    \item[\romannumeral2)] \textbf{Whole-Body Policy}. Instead of only control lower-body joints,
    the whole-body policy controls all 27 joints.
    We train this policy based on the rewards and methods in Exbody \cite{cheng2024expressivewholebodycontrolhumanoid} and HumanPlus\cite{fu2024humanplus}.
    We use this policy to track upper-body motions and keep lower-body joints in default angles.
    \item[\romannumeral3)] \textbf{Decoupled Imitation Policy}. Our main RL policy, which controls lower-body 
    joints to keep balance while tracking upper-body motions.
    \item[\romannumeral4)] \textbf{Decoupled Imitation Policy with Predictive Motion Prior (PMP)\cite{lu2025mobiletelevisionpredictivemotionpriors}}. We add PMP features into the observation state of decoupled policy.
    \item[\romannumeral5)] \textbf{Decoupled Imitation Policy with EMP}. Our full system, RL policy with executable motion prior.
    \item[\romannumeral6)] \textbf{EMP when Danger}. Enable EMP only when regloss of the latent space exceeds 0.04. The regloss reflects the degree of the motion deviation from prior distribution.
    
\end{itemize}

\noindent \textbf{Metrics}.
The metrics are as follows: 
\begin{itemize}
    \item[-]\textbf{Success Rate} (SUC). We define imitation failed when termination conditions in
    section \ref{Termination} are triggered.
    \item[-]\textbf{Mean upper-joint position reward} (MJP). We define upper-body joint 
    position reward as $r_{jp}=\exp{(-\Vert \bm{q}_t - \bm{g}_t \Vert_2)}$.
    \item[-]\textbf{Mean self-collision reward} (MSC). Self-collision often happens while
    tracking motions, which will seriously disturb the balance control of the robots.
    We use link contact force to evaluate this metric: $r_{\mbox{col}}=-\Vert \bm{f}_t \Vert_2 $.
    We only consider the contact between these links: torso, thigh, hand and sacrum.
    \item[-]\textbf{Mean Base Velocity reward} (MBV).
    \item[-]\textbf{Mean Base Acceleration reward} (MBA).
    \item[-]\textbf{Mean Base Orientation reward} (MBO).
    \item[-]\textbf{Mean upper-body action Smoothness} (MUS). We calculate the velocity rate of upper-body joints to evaluate smoothness: $r_{smo} = \Vert \dot{q}_t - \dot{q}_{t-1}\Vert_2^2$
\end{itemize}

\begin{figure}[ht]
    \centering
    \includegraphics[scale=0.35]{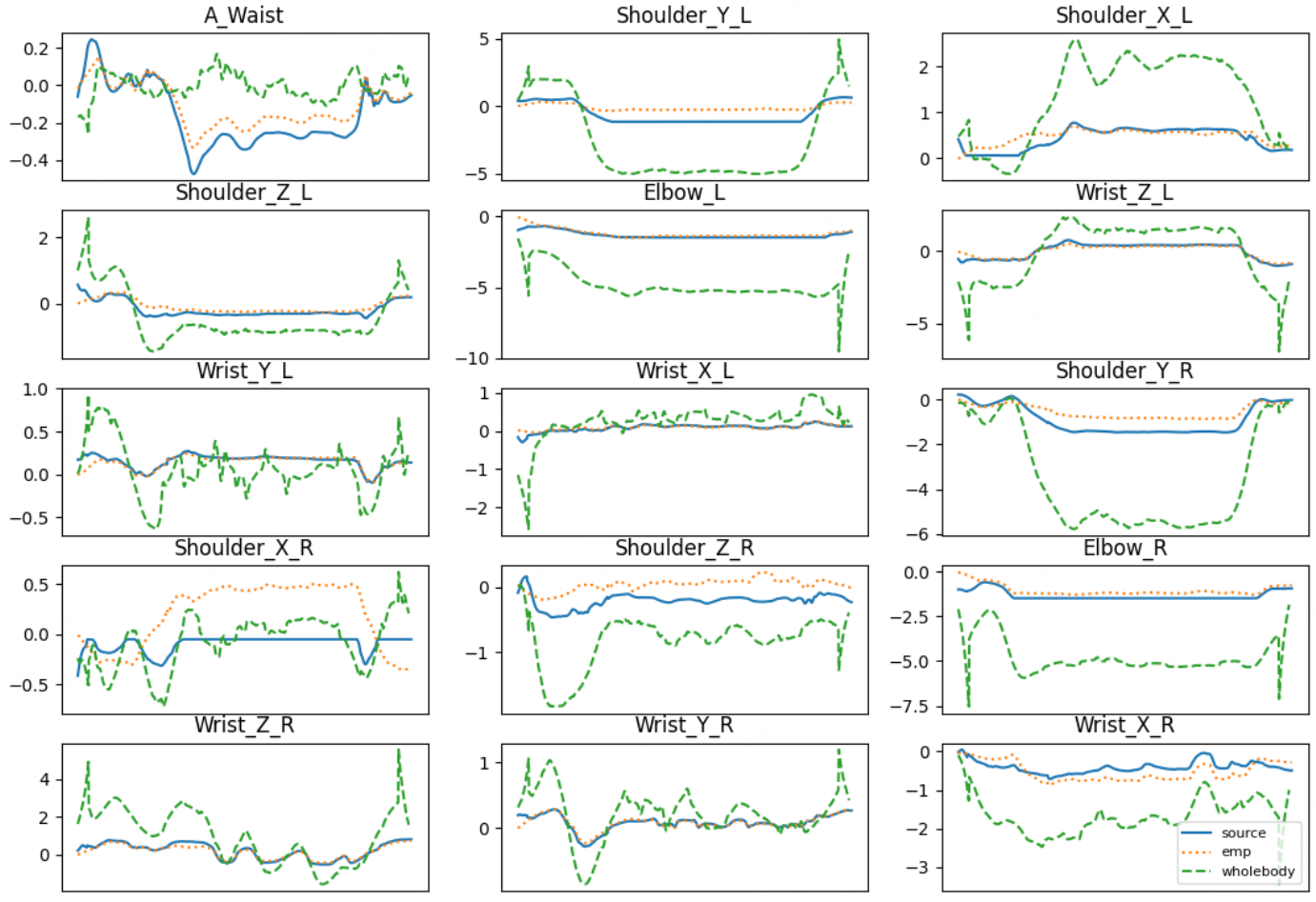}
    \caption{The upper motion (lightbulb screw) of whole-body policy and EMP. The whole-body policy brings significant vibration to upper-body motions.}
    \label{motion compare}
    \vspace{-2.0em}
\end{figure}

\noindent \textbf{Results}.
We deployed our system on the humanoid robot. The results of simulation experiments are summarized in Tab \ref{results naviai}.
Note that we randomly add $0.8 \sim 1.2$kg to both hands as 
hand load. The results reveal that our EMP methods outperform other baselines. Compared with Decoupled Policy, the Whole-Body Policy has a higher success rate and completely avoids collisions (the main reason is that we introduced upper-body collision penalties during training). However, the whole-body policy performs poor in other metrics, especially upper-body motion smoothness. The PMP baseline has achieved a certain improvement in base decoupled policy, but its effectiveness is weaker than that of EMP. 


The EMP network optimizes upper-body motion while minimizing deviations as much as possible, thereby improving control stability. The acceleration, velocity and orientation stability of the base are improved remarkably and the collision is reduced while the joint position error is lightly increased. 

\begin{table*}[ht]
    \vspace{1.0em}
    \renewcommand{\arraystretch}{1.25}
    \newcolumntype{Y}{>{\centering\arraybackslash}X}
    \begin{center}
    \caption{Experiment Results}
    \label{results naviai}
    \begin{tabularx}{\linewidth}{lYYYYYYY}
    \toprule[0.7pt]
    \multirow{2}{*}{Baselines}  &   \multicolumn{7}{c}{Metrics}             \\ 
     \cmidrule(r){2-8}
                                &   SUC $\uparrow$  & MJP $\uparrow$ & MSC $\downarrow $ & MBV $\uparrow$ & MBA $\uparrow$  & MBO $\uparrow$ & MUS $\downarrow$ \\ 
    \hline
    Privileged Policy            &     100\%    &     0.8121     & 0.3856        & 0.8186      &  0.7158        &    0.7702    & 2.4420    \\
    \cmidrule(r){1-1} \cmidrule(r){2-8} 
    Whole-Body Policy            &     \bf{100\%}    &   0.7915       & \bf{0.0}        & 0.7153       &  0.6801    &    0.5204   & 7.6708     \\  
    Decoupled Policy      &     97.0\%    &  \bf{0.8295}   & 0.3668        & 0.7973       &  0.7533       &    0.6699   &   \bf{2.2842}  \\
    PMP     &  97.4\%   &   0.8289       &   0.3741   &  0.7790    & 0.7295        &  0.6800     &  2.3022   \\
    EMP (Ours)   &     98.1\%    &  0.8221   & 0.1494   & \bf{0.8036}  &  0.7588      &    \bf{0.6892}    &    2.3678   \\
    EMP when Danger              &     98.1\%    &    0.8221      & 0.1476        & 0.8029       &  \bf{0.7602}       &    0.6868   & 2.3527    \\
    \toprule[0.7pt]
    \end{tabularx} 
    \end{center}
    \vspace{-2.0em}
\end{table*}

Figure \ref{hi sim} shows some simulation results. While the upper-body motions are executable, our framework maintains consistency with the initial motions. Once the amplitude of the motion exceeds the control capability of the controller, the EMP will optimize the motion to keep the overall robot stable and avoid falling situations.
We analyze the upper-body motion variation curve in Figure \ref{motion compare}. We can see that while acting upper-body motions, whole-body policy exhibits noticeable oscillations, especially in wrist joints.

\begin{table}[h]
    \vspace{-0.5em}
    \renewcommand{\arraystretch}{1.25}
    \newcolumntype{Y}{>{\centering\arraybackslash}X}
    \begin{center}
    \caption{Ablation Study}
    \label{Ablation}
    \begin{tabularx}{\linewidth}{lYYYYYYY}
    \toprule[0.7pt]
    \multirow{2}{*}{Methods}           
                                &   SUC $\uparrow$  & MJP $\uparrow$ & MSC $\downarrow $ & MBV $\uparrow$ & MBA $\uparrow$  & MBO $\uparrow$ & MUS $\downarrow$\\ 
    \hline
    
    Full EMP           &     \bf{98.1\%}    &  \bf{0.822}   & \bf{0.149}   & \bf{0.804}  &  \bf{0.759}     &    \bf{0.689}    &    \bf{2.368}      \\  
    EMP w/o smoothness   &    27.0\%  & 0.637  & 0.211  & 0.702 & 0.591  & 0.555  & 5.434      \\
    EMP w/o orientation   &   2.6\%   & 0.327 & 3.982 & 0.470 & 0.283 & 0.375 & 12.82      \\
    EMP w/o centroid              &   10.7\% & 0.396 & 2.850 &0.531 & 0.232  & 0.422 & 11.00  \\
    
    \toprule[0.7pt]
    \end{tabularx} 
    \end{center}
    \vspace{-2.0em}
    \end{table}
    
\subsection{Ablation Study}

To validate the impact of different losses on the effectiveness of EMP, we conducted ablation experiments on smoothness loss, orientation loss and centroid loss. As illustrated in Tab \ref{Ablation}, the results of the ablation experiments indicate that all three loss functions play an important role in the training of the EMP network. The absence of these loss functions not only affects the directly related metrics but also impacts the overall stability of the system. In contrast, the impact of the smoothness loss on the system is smaller than that of the other two losses.




\subsection{Real-world Experiments}

\noindent \textbf{Deployment Settings}. We test our system on real-world humanoid robot platform.
All the proprioception of the robot comes from build-in sensors. The algorithm we deployed on the real-world system is baseline iv).
Our RL policy and EMP runs at 50Hz. The PD controller is running at 1kHz.

\begin{figure}[h]
    \centering
    \includegraphics[scale=0.35]{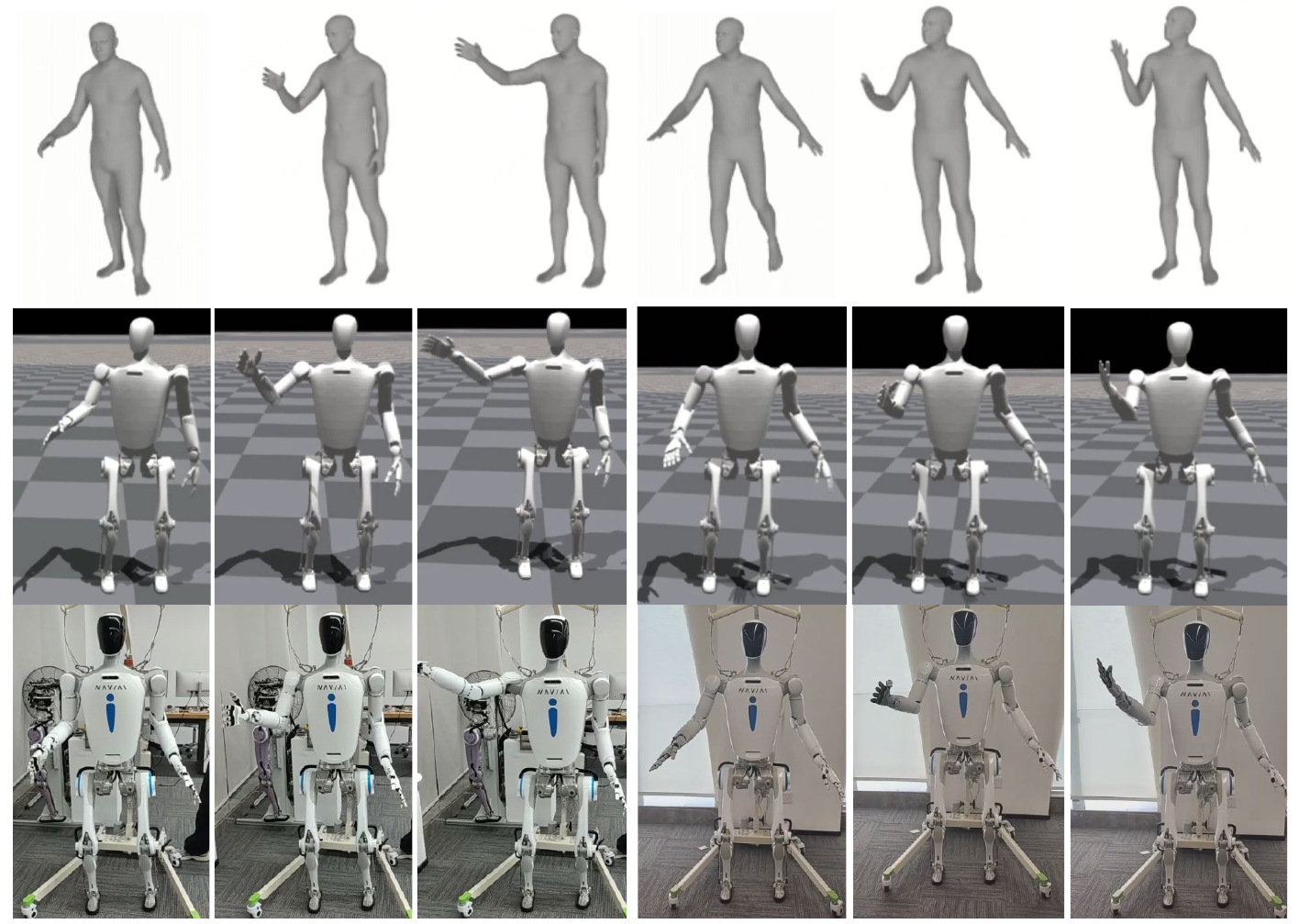}
    \caption{Humanoid robot imitating dataset motion.}
    \label{naviai}
    \vspace{-1.0em}
\end{figure}

\noindent \textbf{Motions Imitation}.
We test several human motions from AMASS dataset in Figure \ref{naviai}. Note that the safety rope connected to the head of the humanoid
is just for protection.

\begin{table}[h]
    \vspace{1.0em}
    \renewcommand{\arraystretch}{1.25}
    \newcolumntype{Y}{>{\centering\arraybackslash}X}
    \begin{center}
    \caption{Simulation Results on Another Robot}
    \label{results navalpha}
    \begin{tabularx}{\linewidth}{lYYYYYYY}
    \toprule[0.7pt]
    \multirow{2}{*}{Baselines}  
                                &   SUC $\uparrow$  & MJP $\uparrow$ & MSC $\downarrow $ & MBV $\uparrow$ & MBA $\uparrow$  & MBO $\uparrow$ & MUS $\downarrow$\\ 
    \hline
    Privileged Policy           &     99.3\%    &   0.840       & 1.317        & 0.787       &  0.284        &    0.702  &  3.642      \\
    \cmidrule(r){1-1} \cmidrule(r){2-8} 
    Whole-body Policy           &     64.8\%    &   0.318       & 2.893        & 0.525       &  0.176    &    0.480   & 8.148      \\  
    Decoupled Policy    &     90.0\%    &   0.841  & 1.748        & 0.792       &  0.381      &    0.727   & 1.604     \\
    EMP (Ours)   &     \bf{97.8\%}    &   \bf{0.861}  & \bf{0.129}   & \bf{0.807}  &  \bf{0.394}      &    \bf{0.754}   & \bf{1.435}       \\
    EMP when Danger              &     95.9\%    &   0.835       & 0.799        & 0.797       &  0.371     &    0.742  & 1.691     \\
    
    \toprule[0.7pt]
    \end{tabularx} 
    \end{center}
    \vspace{-2.0em}
    \end{table}
\subsection{Experiments on Another Platform}
We have also deployed our system on another humanoid platform, which also features two 7-dof arms , two 6-dof legs and
one 1-dof in waist.

\begin{figure}[ht]
    \centering
    \includegraphics[scale=0.45]{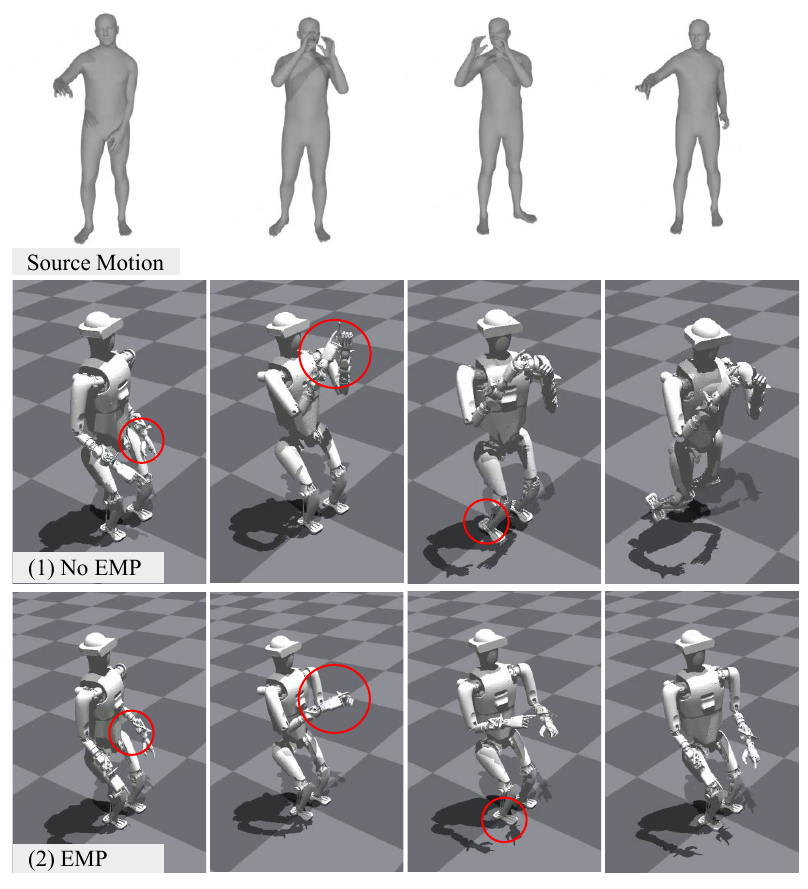}
    \label{binoculars}
    \caption{Simulation experiment on another platform}
    \label{simulation experiments navalpha}
    \vspace{-1.0em}
\end{figure}

The results are shown in Figure \ref{simulation experiments navalpha}, and the metrics of partial baselines are shown in Table \ref{results navalpha}. The results show that our framework also performs well in older platforms.

\section{Conclusions and Future Work}
In this work, we introduce a framework that enables the 
humanoid to imitate upper-body motions retargeted from human motions. 
We train a retargeting network from a humanoid motion dataset and a upper-body
imitation RL policy to control the humanoid to keep balance while tracking motions.
Then our approach utilize executable motion prior before RL controller to transform
difficult motions into executable targets that fit the humanoid control ability.
Through simulations and real-world tests, we validated the effectiveness of our framework. However, we have not realized whole-body motion imitation due
to high DoF and complex dynamics of the full-sized humanoid robot. Meanwhile, joint 
limitations of the robot result in a significant disparity between 
the retargeted motions and the source movements.
We hope to address these limitations in future to build a whole-body
motion imitation system.







\bibliographystyle{IEEEtran}
\bibliography{reference}

\begin{thebibliography}{10}
\providecommand{\url}[1]{#1}
\csname url@rmstyle\endcsname
\providecommand{\newblock}{\relax}
\providecommand{\bibinfo}[2]{#2}
\providecommand\BIBentrySTDinterwordspacing{\spaceskip=0pt\relax}
\providecommand\BIBentryALTinterwordstretchfactor{4}
\providecommand\BIBentryALTinterwordspacing{\spaceskip=\fontdimen2\font plus
\BIBentryALTinterwordstretchfactor\fontdimen3\font minus \fontdimen4\font\relax}
\providecommand\BIBforeignlanguage[2]{{%
\expandafter\ifx\csname l@#1\endcsname\relax
\typeout{** WARNING: IEEEtran.bst: No hyphenation pattern has been}%
\typeout{** loaded for the language `#1'. Using the pattern for}%
\typeout{** the default language instead.}%
\else
\language=\csname l@#1\endcsname
\fi
#2}}

\bibitem{9158331}
Y.~Ishiguro, T.~Makabe, Y.~Nagamatsu, Y.~Kojio, K.~Kojima, F.~Sugai, Y.~Kakiuchi, K.~Okada, and M.~Inaba, ``Bilateral humanoid teleoperation system using whole-body exoskeleton cockpit tablis,'' \emph{IEEE Robotics and Automation Letters}, vol.~5, no.~4, pp. 6419--6426, 2020.

\bibitem{8461207}
Y.~Ishiguro, K.~Kojima, F.~Sugai, S.~Nozawa, Y.~Kakiuchi, K.~Okada, and M.~Inaba, ``High speed whole body dynamic motion experiment with real time master-slave humanoid robot system,'' in \emph{2018 IEEE International Conference on Robotics and Automation (ICRA)}, 2018, pp. 5835--5841.

\bibitem{8375643}
J.~Ramos and S.~Kim, ``Humanoid dynamic synchronization through whole-body bilateral feedback teleoperation,'' \emph{IEEE Transactions on Robotics}, vol.~34, no.~4, pp. 953--965, 2018.

\bibitem{ASE}
\BIBentryALTinterwordspacing
X.~B. Peng, Y.~Guo, L.~Halper, S.~Levine, and S.~Fidler, ``Ase: large-scale reusable adversarial skill embeddings for physically simulated characters,'' \emph{ACM Transactions on Graphics}, vol.~41, no.~4, p. 1–17, July 2022. [Online]. Available: \url{http://dx.doi.org/10.1145/3528223.3530110}
\BIBentrySTDinterwordspacing

\bibitem{AMP}
\BIBentryALTinterwordspacing
X.~B. Peng, Z.~Ma, P.~Abbeel, S.~Levine, and A.~Kanazawa, ``Amp: adversarial motion priors for stylized physics-based character control,'' \emph{ACM Transactions on Graphics}, vol.~40, no.~4, p. 1–20, July 2021. [Online]. Available: \url{http://dx.doi.org/10.1145/3450626.3459670}
\BIBentrySTDinterwordspacing

\bibitem{siekmann2021blindbipedalstairtraversal}
\BIBentryALTinterwordspacing
J.~Siekmann, K.~Green, J.~Warila, A.~Fern, and J.~Hurst, ``Blind bipedal stair traversal via sim-to-real reinforcement learning,'' 2021. [Online]. Available: \url{https://arxiv.org/abs/2105.08328}
\BIBentrySTDinterwordspacing

\bibitem{li2024reinforcementlearningversatiledynamic}
\BIBentryALTinterwordspacing
Z.~Li, X.~B. Peng, P.~Abbeel, S.~Levine, G.~Berseth, and K.~Sreenath, ``Reinforcement learning for versatile, dynamic, and robust bipedal locomotion control,'' 2024. [Online]. Available: \url{https://arxiv.org/abs/2401.16889}
\BIBentrySTDinterwordspacing

\bibitem{cheng2024expressivewholebodycontrolhumanoid}
\BIBentryALTinterwordspacing
X.~Cheng, Y.~Ji, J.~Chen, R.~Yang, G.~Yang, and X.~Wang, ``Expressive whole-body control for humanoid robots,'' 2024. [Online]. Available: \url{https://arxiv.org/abs/2402.16796}
\BIBentrySTDinterwordspacing

\bibitem{he2024learninghumantohumanoidrealtimewholebody}
\BIBentryALTinterwordspacing
T.~He, Z.~Luo, W.~Xiao, C.~Zhang, K.~Kitani, C.~Liu, and G.~Shi, ``Learning human-to-humanoid real-time whole-body teleoperation,'' 2024. [Online]. Available: \url{https://arxiv.org/abs/2403.04436}
\BIBentrySTDinterwordspacing

\bibitem{lu2025mobiletelevisionpredictivemotionpriors}
\BIBentryALTinterwordspacing
C.~Lu, X.~Cheng, J.~Li, S.~Yang, M.~Ji, C.~Yuan, G.~Yang, S.~Yi, and X.~Wang, ``Mobile-television: Predictive motion priors for humanoid whole-body control,'' 2025. [Online]. Available: \url{https://arxiv.org/abs/2412.07773}
\BIBentrySTDinterwordspacing

\bibitem{fu2024humanplus}
Z.~Fu, Q.~Zhao, Q.~Wu, G.~Wetzstein, and C.~Finn, ``Humanplus: Humanoid shadowing and imitation from humans,'' in \emph{arXiv}, 2024.

\bibitem{10.1145/311535.311536}
\BIBentryALTinterwordspacing
Z.~Popovi\'{c} and A.~Witkin, ``Physically based motion transformation,'' in \emph{Proceedings of the 26th Annual Conference on Computer Graphics and Interactive Techniques}, ser. SIGGRAPH '99.\hskip 1em plus 0.5em minus 0.4em\relax USA: ACM Press/Addison-Wesley Publishing Co., 1999, pp. 11--20. [Online]. Available: \url{https://doi.org/10.1145/311535.311536}
\BIBentrySTDinterwordspacing

\bibitem{rekik2024correspondencefreeonlinehumanmotion}
\BIBentryALTinterwordspacing
R.~Rekik, M.~Marsot, A.-H. Olivier, J.-S. Franco, and S.~Wuhrer, ``Correspondence-free online human motion retargeting,'' 2024. [Online]. Available: \url{https://arxiv.org/abs/2302.00556}
\BIBentrySTDinterwordspacing

\bibitem{10.1145/2897824.2925975}
\BIBentryALTinterwordspacing
D.~Holden, J.~Saito, and T.~Komura, ``A deep learning framework for character motion synthesis and editing,'' \emph{ACM Trans. Graph.}, vol.~35, no.~4, jul 2016. [Online]. Available: \url{https://doi.org/10.1145/2897824.2925975}
\BIBentrySTDinterwordspacing

\bibitem{10.1145/3283289.3283316}
\BIBentryALTinterwordspacing
H.~Jang, B.~Kwon, M.~Yu, S.~U. Kim, and J.~Kim, ``A variational u-net for motion retargeting,'' in \emph{SIGGRAPH Asia 2018 Posters}, ser. SA '18.\hskip 1em plus 0.5em minus 0.4em\relax New York, NY, USA: Association for Computing Machinery, 2018. [Online]. Available: \url{https://doi.org/10.1145/3283289.3283316}
\BIBentrySTDinterwordspacing

\bibitem{Aberman_2020}
\BIBentryALTinterwordspacing
K.~Aberman, P.~Li, D.~Lischinski, O.~Sorkine-Hornung, D.~Cohen-Or, and B.~Chen, ``Skeleton-aware networks for deep motion retargeting,'' \emph{ACM Transactions on Graphics}, vol.~39, no.~4, Aug. 2020. [Online]. Available: \url{http://dx.doi.org/10.1145/3386569.3392462}
\BIBentrySTDinterwordspacing

\bibitem{10496240}
H.~Zhang, Z.~Chen, H.~Xu, L.~Hao, X.~Wu, S.~Xu, R.~Xiong, and Y.~Wang, ``Unified cross-structural motion retargeting for humanoid characters,'' \emph{IEEE Transactions on Visualization and Computer Graphics}, pp. 1--14, 2024.

\bibitem{7966379}
B.~Delhaisse, D.~Esteban, L.~Rozo, and D.~Caldwell, ``Transfer learning of shared latent spaces between robots with similar kinematic structure,'' in \emph{2017 International Joint Conference on Neural Networks (IJCNN)}, 2017, pp. 4142--4149.

\bibitem{8063558}
K.~Ayusawa and E.~Yoshida, ``Motion retargeting for humanoid robots based on simultaneous morphing parameter identification and motion optimization,'' \emph{IEEE Transactions on Robotics}, vol.~33, no.~6, pp. 1343--1357, 2017.

\bibitem{luo2023perpetualhumanoidcontrolrealtime}
\BIBentryALTinterwordspacing
Z.~Luo, J.~Cao, A.~Winkler, K.~Kitani, and W.~Xu, ``Perpetual humanoid control for real-time simulated avatars,'' 2023. [Online]. Available: \url{https://arxiv.org/abs/2305.06456}
\BIBentrySTDinterwordspacing

\bibitem{10.5555/3495724.3497550}
Y.~Yuan and K.~M. Kitani, ``Residual force control for agile human behavior imitation and extended motion synthesis,'' in \emph{Proceedings of the 34th International Conference on Neural Information Processing Systems}, ser. NIPS '20.\hskip 1em plus 0.5em minus 0.4em\relax Red Hook, NY, USA: Curran Associates Inc., 2020.

\bibitem{HumanoidTransformer2023}
I.~Radosavovic, T.~Xiao, B.~Zhang, T.~Darrell, J.~Malik, and K.~Sreenath, ``Learning humanoid locomotion with transformers,'' \emph{arXiv:2303.03381}, 2023.

\bibitem{10035484}
K.~Darvish, L.~Penco, J.~Ramos, R.~Cisneros, J.~Pratt, E.~Yoshida, S.~Ivaldi, and D.~Pucci, ``Teleoperation of humanoid robots: A survey,'' \emph{IEEE Transactions on Robotics}, vol.~39, no.~3, pp. 1706--1727, 2023.

\bibitem{zhang2023slomogeneralleggedrobot}
\BIBentryALTinterwordspacing
J.~Z. Zhang, S.~Yang, G.~Yang, A.~L. Bishop, D.~Ramanan, and Z.~Manchester, ``Slomo: A general system for legged robot motion imitation from casual videos,'' 2023. [Online]. Available: \url{https://arxiv.org/abs/2304.14389}
\BIBentrySTDinterwordspacing

\bibitem{doi:10.1126/scirobotics.aav4282}
\BIBentryALTinterwordspacing
J.~Ramos and S.~Kim, ``Dynamic locomotion synchronization of bipedal robot and human operator via bilateral feedback teleoperation,'' \emph{Science Robotics}, vol.~4, no.~35, p. eaav4282, 2019. [Online]. Available: \url{https://www.science.org/doi/abs/10.1126/scirobotics.aav4282}
\BIBentrySTDinterwordspacing

\bibitem{oord2018neuraldiscreterepresentationlearning}
\BIBentryALTinterwordspacing
A.~van~den Oord, O.~Vinyals, and K.~Kavukcuoglu, ``Neural discrete representation learning,'' 2018. [Online]. Available: \url{https://arxiv.org/abs/1711.00937}
\BIBentrySTDinterwordspacing

\bibitem{zhang2022kinematicmotionretargetingneural}
\BIBentryALTinterwordspacing
H.~Zhang, W.~Li, J.~Liu, Z.~Chen, Y.~Cui, Y.~Wang, and R.~Xiong, ``Kinematic motion retargeting via neural latent optimization for learning sign language,'' 2022. [Online]. Available: \url{https://arxiv.org/abs/2103.08882}
\BIBentrySTDinterwordspacing

\bibitem{kingma2022autoencodingvariationalbayes}
\BIBentryALTinterwordspacing
D.~P. Kingma and M.~Welling, ``Auto-encoding variational bayes,'' 2022. [Online]. Available: \url{https://arxiv.org/abs/1312.6114}
\BIBentrySTDinterwordspacing

\bibitem{shao2020controlvaecontrollablevariationalautoencoder}
\BIBentryALTinterwordspacing
H.~Shao, S.~Yao, D.~Sun, A.~Zhang, S.~Liu, D.~Liu, J.~Wang, and T.~Abdelzaher, ``Controlvae: Controllable variational autoencoder,'' 2020. [Online]. Available: \url{https://arxiv.org/abs/2004.05988}
\BIBentrySTDinterwordspacing

\bibitem{makoviychuk2021isaacgymhighperformance}
\BIBentryALTinterwordspacing
V.~Makoviychuk, L.~Wawrzyniak, Y.~Guo, M.~Lu, K.~Storey, M.~Macklin, D.~Hoeller, N.~Rudin, A.~Allshire, A.~Handa, and G.~State, ``Isaac gym: High performance gpu-based physics simulation for robot learning,'' 2021. [Online]. Available: \url{https://arxiv.org/abs/2108.10470}
\BIBentrySTDinterwordspacing

\bibitem{Taheri_2020}
\BIBentryALTinterwordspacing
O.~Taheri, N.~Ghorbani, M.~J. Black, and D.~Tzionas, \emph{GRAB: A Dataset of Whole-Body Human Grasping of Objects}.\hskip 1em plus 0.5em minus 0.4em\relax Springer International Publishing, 2020, pp. 581--600. [Online]. Available: \url{http://dx.doi.org/10.1007/978-3-030-58548-8_34}
\BIBentrySTDinterwordspacing

\bibitem{AMASS:ICCV:2019}
N.~Mahmood, N.~Ghorbani, N.~F. Troje, G.~Pons-Moll, and M.~J. Black, ``{AMASS}: Archive of motion capture as surface shapes,'' in \emph{International Conference on Computer Vision}, Oct. 2019, pp. 5442--5451.

\bibitem{kingma2017adammethodstochasticoptimization}
\BIBentryALTinterwordspacing
D.~P. Kingma and J.~Ba, ``Adam: A method for stochastic optimization,'' 2017. [Online]. Available: \url{https://arxiv.org/abs/1412.6980}
\BIBentrySTDinterwordspacing

\end{thebibliography}

\end{document}